\documentclass[10pt,twocolumn,letterpaper]{article}
\usepackage{cvpr}      

\usepackage{graphicx}
\usepackage{amsmath}
\usepackage{amssymb}
\usepackage{booktabs}

\usepackage[pagebackref,breaklinks,colorlinks]{hyperref}
\usepackage[capitalize]{cleveref}
\crefname{section}{Sec.}{Secs.}
\Crefname{section}{Section}{Sections}
\Crefname{table}{Table}{Tables}
\crefname{table}{Tab.}{Tabs.}

\def\cvprPaperID{*****} 
\def\confName{CVPR}
\def\confYear{2022}

\usepackage{gensymb}
\usepackage[usenames,dvipsnames]{xcolor}
\usepackage{color}
\usepackage[linesnumbered,titlenumbered,ruled,vlined,resetcount,algosection]{algorithm2e}
\usepackage[OT1]{fontenc} 
\begin{document}

\title{RLM-Tracking: Online Multi-Pedestrian Tracking Supported by Relative Location Mapping}

\author{
Kai Ren,
~~~    
Chuanping Hu
\\[0.2cm] 
The University of Zhengzhou \\
}
\maketitle

\begin{abstract}
  The problem of multi-object tracking is a fundamental computer vision research focus, widely used in public safety, transport, autonomous vehicles, robotics, and other regions involving artificial intelligence. Because of the complexity of natural scenes, object occlusion and semi-occlusion usually occur in fundamental tracking tasks. These can easily lead to ID switching, object loss, detect errors, and misaligned limitation boxes. These conditions have a significant impact on the precision of multi-object tracking. In this paper, we design a new multi-object tracker for the above issues that contains an object \textbf{Relative Location Mapping} (RLM) model and \textbf{Target Region Density} (TRD) model. The new tracker is more sensitive to the differences in position relationships between objects. It can introduce low-score detection frames into different regions in real-time according to the density of object regions in the video. This improves the accuracy of object tracking without consuming extensive arithmetic resources. Our study shows that the proposed model has considerably enhanced the HOTA and DF1 measurements on the MOT17 and MOT20 data sets when applied to the advanced MOT method.
\end{abstract}

  \section{Introduction}
\label{sec:Introduction}

Multi-object tracking (MOT) technology is a machine vision task. It has been mainly applied in autonomous vehicles, robot navigation, and public video analysis in recent years. With the rapid development of object detection technology, detection-tracking paradigm has become the most effective multi-object tracking scheme. An increasing number of results show that detection tracking methods can achieve well in multi-object tracking. However, in practical scenarios, inter-target occlusion and out-of-target occlusion issues occurring during the motion of multiple objects have always been the challenges faced by MOT. In MOT scenes, to maintain consistency before and after the same object in the video. The object detection algorithm typically uses the method for re-identifying \cite{Fu2020} (Re-ID) the objects in each image in the video. Given the rapid development of object detection technology, researchers generally use the re-ID model independently of the object detection algorithm alone. In early studies, object re-identification methods usually enter the detected object directly after sliding into the pre-formed Re-ID model and use deep neural network techniques to calculate the features of the object. This significantly increases the calculation cost and requires more parameters to respond to the selection of the object deviation to obtain more apparent internal characteristics. However, the multiobject tracking application is generally characterised by high demands in real-time. To achieve real-time tracking, some studies \cite{Wang2020b} attempt to reduce the repeated calculation of the target image characteristic values by sharing the anchorage or point detector characteristics in the object detection backbone network. Because of the problem of the object’s occlusion, certain studies \cite{felzenszwalb2008discriminatively} try to detect and compare the object by detecting only the visible part of the object, dividing the object into several parts one at a time, or making multiple predictions for a single detection base. Point-based detectors \cite{Zhou2020} solve the problem of target image repetition rates being too high in masking, so low-score images are suppressed by non-axial suppression (NMS).

With the maturation of multi-task learning technology, some research \cite{Li2022,Zhou2022} explores the earlier detection tracking paradigm to a one-time tracker paradigm, such as target estimation and identity re-identification. These studies add Re-ID branches to the backbone network to obtain the features of each target, which is similar to the method of sharing the feature values of the backbone network and makes the overall performance of the scheme significantly lower compared to the two-step detection tracking paradigm, mainly because adding Re-ID branches to the backbone network to memorise targets consumes a lot of computation and storage costs to maintain some identity information that is not necessarily. This leads to system redundancy that affects performance. The one-time tracking anchor approach is unsuitable for the simultaneous extraction of both detection object features and Re-ID features because the types and dimensions of features used are different.

The Re-ID method based on feature values predicts the object trajectory after obtaining the object features combined with the inter-frame relationship. It then uses the association algorithm to correlate the object and trajectory with each other to get the object path. However, as some low-dimensional characteristics may obtain fewer differences in object intraclass features, the video shows the distance between the two-dimensional marks. It is easy to confuse the identities of a similar object and generate switches during object occlusion. When using centroid detectors to detect objects, the relative position relation between the object is not always precisely reflected in the figure. And the factors affecting the centroid position are not just the position of the object but also the object’s distance, the shape of the mark and the positioning of the object. Consequently, errors can occur if we rely solely on the target centroid to calculate the trajectory of the object.

In this paper, we analyze the causes of incorrect tracking generated by previous trackers and offer a simple and efficient auxiliary tracking system with the following main contributions.

1. The relative location mapping model proposed in this document enhances the treatment of differences in object position relations in the monitoring process, effectively reducing the frequency of target changes due to body size, posture and pedestrian occlusion factors;

2. The TRD model proposed here quantifies the target density of different regions of the video image, allowing the tracker to adjust the threshold value of the low-score detection image adaptively according to the size of the target density in different regions, reduce the interference of the low score detection frame with the low-density target area and reduce the computational cost;

3. After using the relative position mapping model to project the object from the video image plane in the mapping plane, we propose adopting the standardised bounding boxes approach to consistently label the object's position in the mapping plane. This can improve the tracker's perception of target movement speed and position changes, thereby improving tracking accuracy;

\begin{figure*}[ht]
  \centering
  \captionsetup{justification=centering}
  \includegraphics[width=1\textwidth]{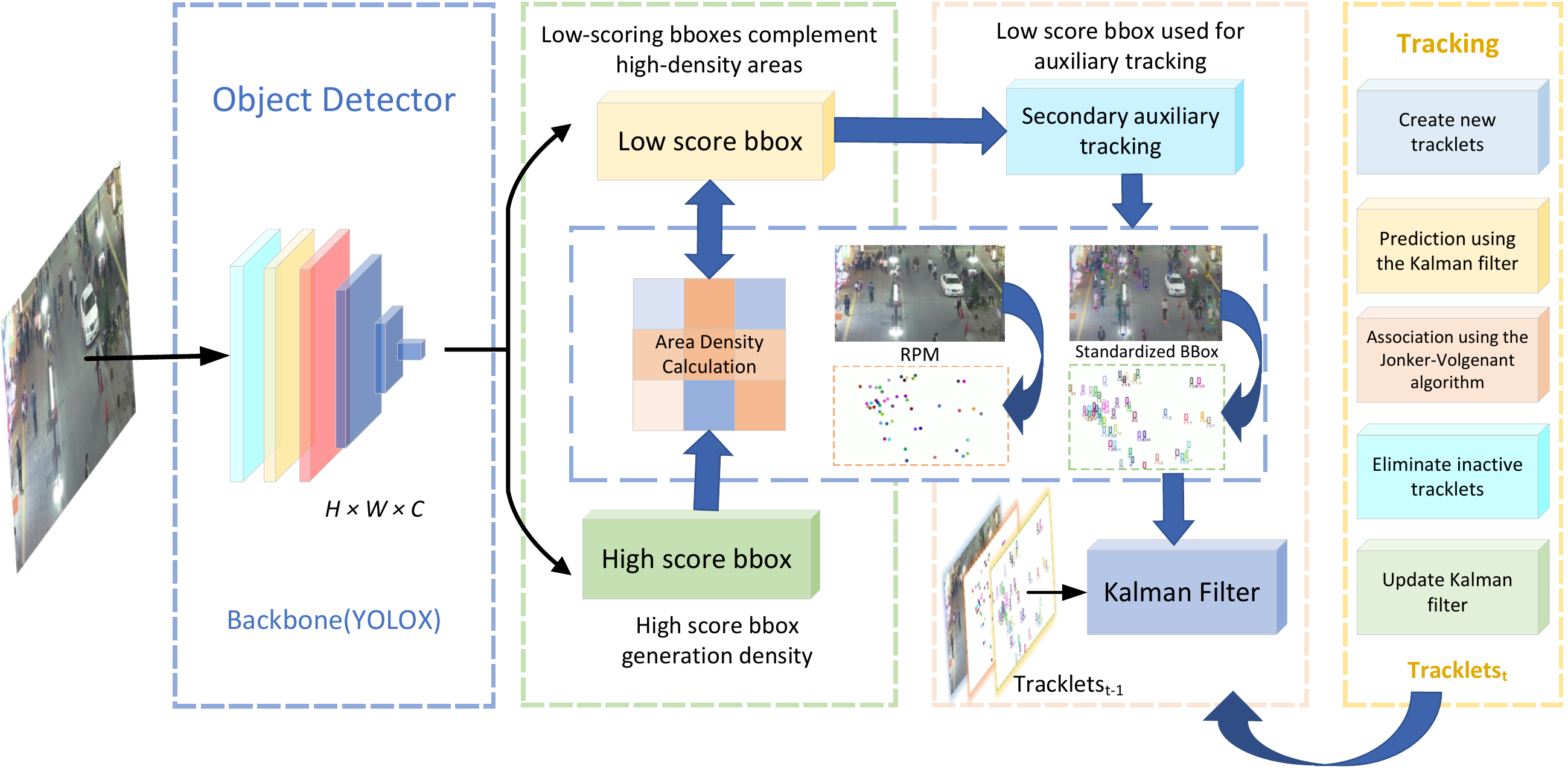}
  \caption{relative position mapping tracking method outline diagram}
  \label{figure:1}
\end{figure*}
\section{Related Work}

There are two major types of current approaches to multiple object tracking research, which are the detection followed by TBD (tracking by detection) paradigm \cite{Zhang2021,Du2022,Wan2022,Zhang2021a,Bergmann2019} and JDT (joint detection and tracking) paradigm \cite{Chu2020a,Saito2021,Reading2021,Husain2021,Sun2021, Kervadec2021}. There are different purposes for these two types of research. The former primarily benefits from the many models developed within the current object detection field, allowing researchers to quickly and accurately obtain object localization normalization suggestion box within video image frames, then correlate the same identity normalization suggestion box together to get the target trajectory by association algorithms; the latter usually integrates the object detection and identity association work into a neural network to obtain more efficient operation and end-to-end tracking.

\subsection{Tracking By Detection Paradigm}

The detection followed by the tracking paradigm is the current multi-object tracking method that runs the detection and tracking independently as two models. Detection generally employs the most advanced or on-demand object detection models, such as FasterRCNN\cite{Ren2017}, YOLO\cite{Redmon2016}, Transformer\cite{Carion2020}, and the latest Optimized versions like Cascade R-CNN\ cite{cai2018cascade}, YOLOX\cite{Ge2021}, DINO\cite{Zhang2022}, swim Transformer\cite{Liu2021}. Object detection is the fundamental study of computer vision, and it includes several detection models, excellent detection speed and accuracy. Researchers using the post-detection monitoring paradigm can put more energy into researching object associations. Still, the computational overload of the post-detection tracking paradigm is the sum of the two. The improvement in real-time tracking performance with the selected object detection model can only be increased by enhancing the correlation model to increase efficiency.

In addition, the TBD paradigm is divided into online and offline tracking depending on the degree of real-time processing of video images. Online tracking \cite{Cao2022, Peng2020, Zheng2021, Zhou2020, Chu2017, Zhou2018, Zhu2018} is according to current and previous video images to follow the path of the object. In contrast, offline tracking \cite{Kim2015, He2022} is more for all frames or batch frames of a specific off-line video to resolve the object tracking issue. Since multi-object tracking is mainly used in practical engineering applications like self-driving, monitoring and robotics, real-time is an essential measurement, and processing of localized video images will reveal faster processing effectiveness. The literature \cite{Bewley2016} using the SORT study proposes the use of Kalman filter \cite{welch1995introduction,ren2021kalman} to predict the future object position according to the evolution of the object's trajectory in the preceding images. Then the Hungarian algorithm \cite{Kuhn1955} to correlate tracking results, simple and effective as its main feature, but easy to produce loss and misjudgment for obscured targets. The DeepSORT of literature \cite{Wojke2018} adds the target appearance depth information to enhance the effectiveness of association based on the study of SORT, which can reduce the occlusion effect on the tracking effect. The study of literature \cite{He2022} proposes an identity number harmonic tracking framework IQHAT, which allows multiple target identities to correspond to multiple targets, and the scheme performs better for task occlusion in crowded scenes. The StrongSORT study of literature \cite{Du2022} is based on an upgraded version of DeepSORT, which proposes an appearance-free linking model to correlate short and complete trajectories and compensate for missing detection using Gaussian smoothing differences, and obtains better results. The study \cite{Zhang2021} proposes a method to balance detection and association weights and analyzes that shallow target features are more accessible to obtain standard features of the same target than in-depth features. The study \cite{Zhang2021a} also proposes that when dealing with low-score target objects generated during detection, the low-score detection should be judged again carefully to confirm whether it is an occluded target. The experiment proves that this approach can significantly improve the accuracy of tracking.
\begin{figure*}[ht]
  \centering
  \captionsetup{justification=centering}
  \includegraphics[width=0.90\textwidth]{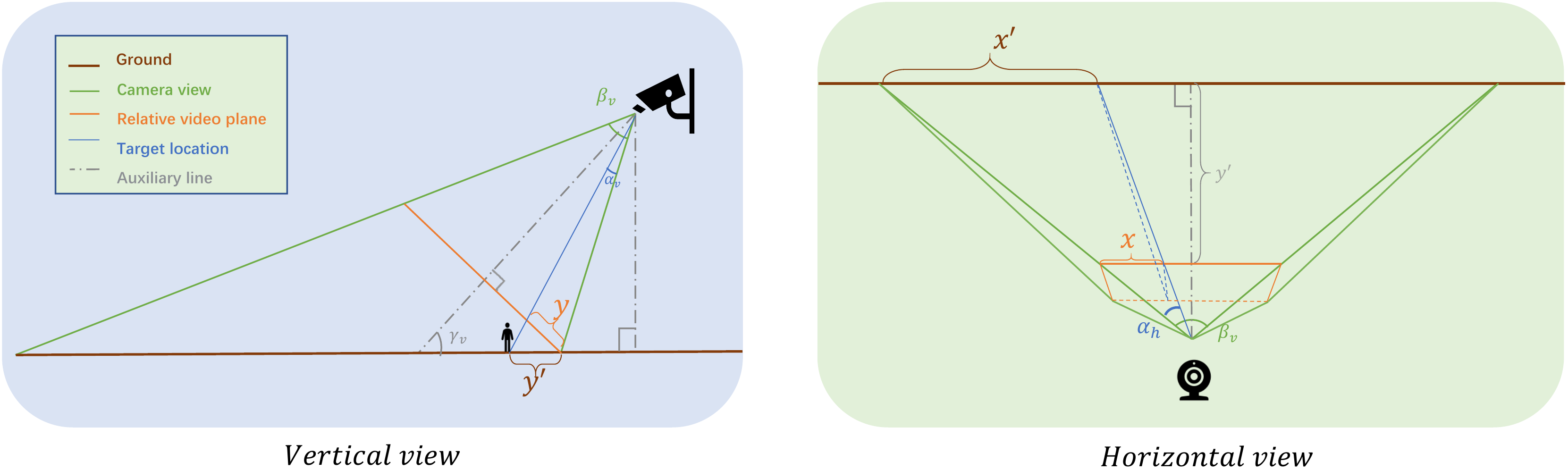}
  \caption{vertical and horizontal view sketch map}
  \label{figure:2}
\end{figure*}
\subsection{Joint Detection and Tracking Paradigm}

In addition to the inclusion of traditional CNN approaches, thanks to the breakthrough development in the field of Natural language processing (NLP), single-model architectures based on transformers have started to be increasingly studied in the image domain. In this paper, \cite{Wang2020b} designed a JDE method to combine object detection and ID association in the same model with shared parameters, improving the multi-object detection efficiency while extracting object association features through deep learning and reducing the impact of occlusion to a certain extent. The literature \cite{Zeng2021} extends DETR using the MOTR method and models the temporal relationships of temporal changes in video sequences to improve tracking accuracy. The literature \cite{Liang2022} uses a study of CSTrack to analyze the performance problems caused by the existence of neglected discrepancy factors in the processing of detection and association in a single model and proposes interaction network REN as scale aware attention network SAAN. The study of literature \cite{Zhou2020} using CenterTrack adopts the center point position heat map of the object detection frame and the previous frame RGB map as the object tracking point track instead of the object detection frame, which can reduce the complexity of the bounding box approach to detection tracking and get the best balance in efficiency and accuracy. The studies in the literature \cite{Xu2021, Sun2020, Wan2022, Yu2022} use an attention mechanism to enhance the connection between the preceding and following frames to improve the association efficiency. Literature \cite{Wu2021c} analyzed that the reason for the lower effectiveness of single-model Multi-object tracking compared to the dual-model structure is due to the difference in the feature metrics required for object detection and object association, where the requirement for object detection is a sizeable inter-class feature differentiation and a little intra-class differentiation. In contrast, object association needs a sizeable intra-class differentiation, so a single model is not good enough to satisfy both requirements.
\subsection{This Paper}
In earlier multi-object tracking studies, partial detection or trajectory prediction is used directly in the image when the tracking target is occlusive. Nonetheless, target overlap and occlusion can occur due to the image capture angle, resulting in frequent identity changes and target losses during follow-up. In this paper, we use the TBD paradigm to map the target location obtained by the detector in various scenes and project the target in the video picture onto the ground map according to a specific relation. We use YOLOX\cite{Ge2021} trained through additional datasets in the object detection module and the associative follow-up idea with low-score bounding boxes in ByteTrack\cite{Zhang2021a}. We are also rethinking how standard target bounding boxes are generated in the mapping map. We propose a TRD model to quantify regions where video images are likely occlusive. More adaptively adjust the size of the normalization suggestion box on the map and the threshold of low-scoring target bounding boxes by occlusion probability. Finally, we also use the TRD model to adjust the Kalman gain coefficients in real-time to increase tracking stability. This improvement is experimentally demonstrated to reduce ID switching and target loss probabilities from tracking objects in congested scenes.

\section{Relative Position Mapping}
In a video image, the shooter will shoot pedestrians on the road at a certain angle. The lens faces the crowd or the dense human traffic zone for a complete picture of the target scene. However, the problem is that the target pedestrian image will be superimposed on the image to occur occlusive phenomenon. Therefore, projecting the object's position in the video image on the ground map is especially important to address the problem of tracking the object.
\subsection{Relative Position Mapping Tracking}
\begin{figure}[ht]
  \centering
  \captionsetup{justification=centering}
  \includegraphics[width=0.45\textwidth]{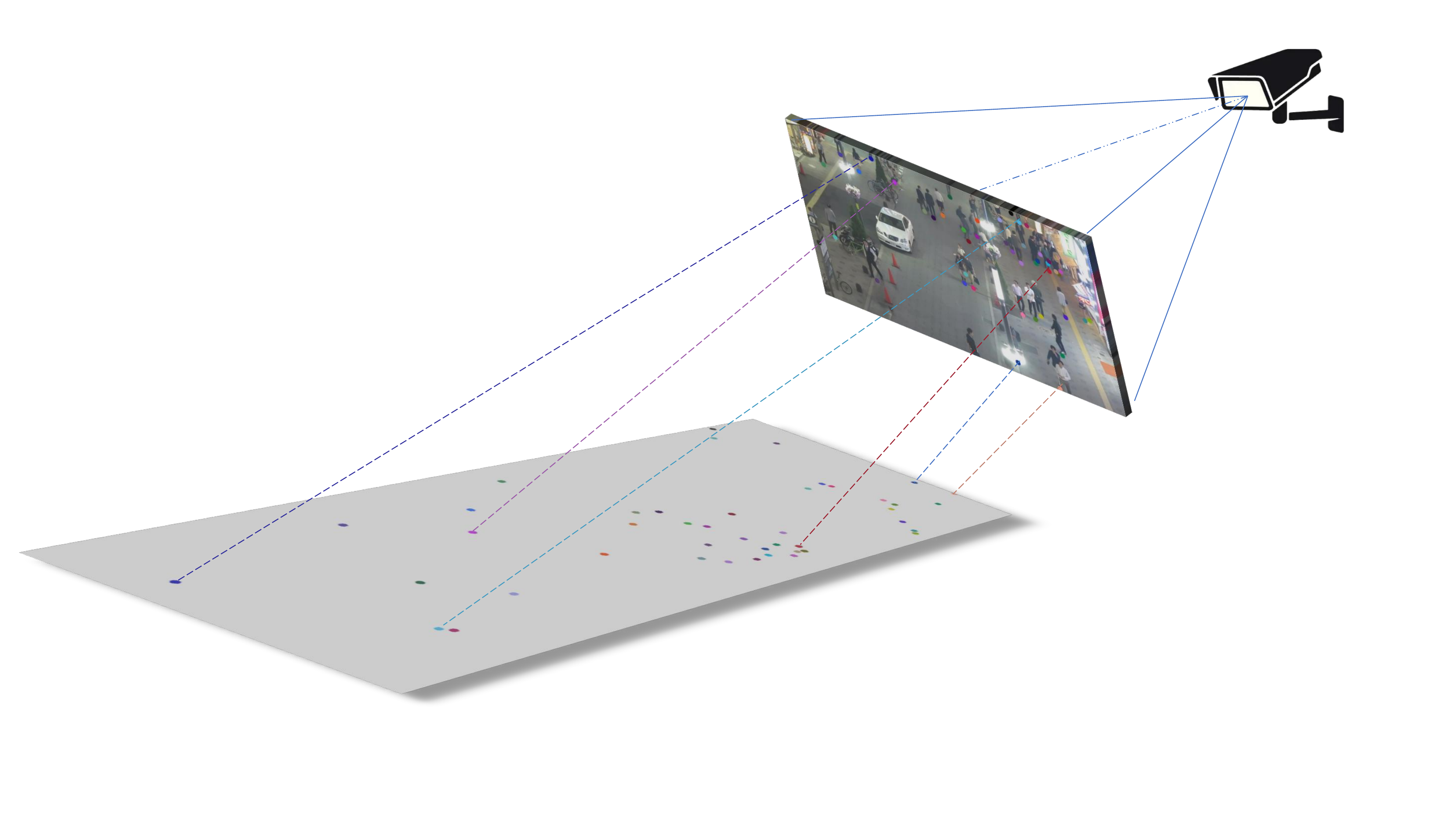}
  \caption{Relative location mapping (RLM)}
  \label{figure:3}
\end{figure}

We propose a simple and effective Relative Location Mapping (RLM) model, schematically shown in \ref{figure:3}, which reduces the position of the video image on the mapping map based on the optical imaging features.

\begin{algorithm}[!ht]
  \DontPrintSemicolon
  \SetAlgoLined
  \SetNoFillComment
  \KwIn{A video sequence $V$; object detector $Det$; region density matrix $A_\rho $; Relative location mapping $RLM$; detection score threshold $\tau$ ; standardized bounding box $\mathcal{F} $}
  \KwOut {Tracks $\mathcal{T} $ of the video}
  initialization:
  $\mathcal{T} \gets \emptyset $\;
  \For{frame $f_k$ in $V$}{
  \tcc*[h]{Figure \ref{figure:1}}\;
  \tcc*[h]{detection bboxes \& scores }\;
  $\mathcal{D}_k \gets \emptyset $\;
  $\mathcal{D}_{high} \gets \emptyset $\;
  $\mathcal{D}_{low} \gets \emptyset $\;
  $\mathcal{A}_\rho  \gets [\emptyset] $\;

  \For{bbox  $b$, score  in $\mathcal{D}_k$ }{
    \If{$score > \tau_{high}$ }{
      $\mathcal{D} _{high} \gets \mathcal{D} _{high}\cup \{b\}$\;
      \tcc*[h]{region density matrix}
      $\color{ForestGreen}{\mathcal{A} _\rho \gets \mathcal{F} (\mathcal{D} _{high})}$

    }\ElseIf{$score > \color{ForestGreen}{\mathcal{A} _\rho * \tau_{low}  }$}{
      $\mathcal{D} _{low} \gets \mathcal{D} _{low} \cup \{b\}$
    }
  }
  \tcc*[h]{Relative location mapping and standard bounding boxes}\;
  $\color{ForestGreen}{\mathcal{R} _{high},\mathcal{R} _{low}  \gets RLM(\mathcal{D} _{high},\mathcal{D} _{low})}$\;
  $\color{ForestGreen}{\mathcal{D}'_{high}, \mathcal{D}'_{low} \gets \mathcal{F}(\mathcal{R} _{high},\mathcal{R} _{low},\mathcal{A} _\rho)  }$\;
  \tcc*[h]{predict with region density by Kalmanfilter after the mapping}\;
  \For{$t' \:in\: \mathcal{T}' $}{
    $\color{ForestGreen}{t' \gets Kalmanfilter(t',\mathcal{A}_{\rho })}$\;
  }
  \tcc*[h]{first association}\;
  Associate $\mathcal{T}'$ and $\mathcal{D}'_{high}$ using IoU\;
  $\mathcal{D}'_{useless} \gets  (\mathcal{D}'_{high} - \mathcal{D}'_{used})$\;
  $\mathcal{T}'_{unmatched1} \gets  (\mathcal{T}' - \mathcal{D}'_{tracked})$\;
  \tcc*[h]{second association}\;
  Associate $\mathcal{T}'_{unmatched1}$ and $\mathcal{D}'_{low}$ using IoU\;
  $\mathcal{T}'_{unmatched2} \gets  (\mathcal{T}'_{unmatched1} - \mathcal{D}'_{tracked})$\;
  \tcc*[h]{delete unmatched tracks and add new tracks}\;
  $\mathcal{T}' \gets (\mathcal{T}' - \mathcal{T}'_{unmatched2} + \mathcal{D}'_{useless})$
  \tcc*[h]{restore the original bbox}\;
  \For(){$t'\: in\: \mathcal{T}'$}{
    $\color{ForestGreen}{\mathcal{T} \gets RLM'(t')}$
  }
  }
  Return: $\mathcal{T} $
  \caption{Pseudo-code of RLMTrack}
  \label{Algorithm:1}
\end{algorithm}
The algorithm \ref{Algorithm:1} shows the full process of the tracker using this model to handle the video image tracking task. Our algorithm continues the idea of \cite{Zhang2021} to introduce a low-scoring bounding box and is different from the previous \cite{Wojke2018, Bewley2016, Wang2021, Cao2022} that associates the target bounding box directly for computation. The algorithm is divided into five stages: detection frame screening, relative position mapping, initial tracker association, secondary tracker association, and tracker output result. After the detector detects the video image frame by frame, this algorithm first selects the high-scoring bounding box with a score greater than $\tau_{high} $ to sink the object position coordinate weight to the bottom center of the bounding box. The tracker calculates the region density matrix $A_\rho $ according to the TRD. After getting the region density matrix, the algorithm filters out the low-scoring bounding box with scores greater than $A_\rho * \tau_{low} $ according to the density of different areas to participate in the supplementary tracking association. Then the object position coordinates obtained above are converted to the mapping plane using the $RLM$ model to obtain the relative position coordinates $\mathcal{R} _{high},\mathcal{R} _{low} $, and then the complete normalized suggestion box $\mathcal{D} _{high},\ mathcal{D} _{low} $. At this point, the tracker completes the relative position mapping conversion task, and we put the converted bounding box parameters and their region density values into the Kalman filter algorithm to calculate the predicted position of the target in the mapping plane for the next frame. The tracker performs an association operation on the object position of each frame. It will first associate the high-score bounding box and then use the low-score bounding box to complement the tracking task to obtain $\mathcal{T}' $. Finally, it also needs to perform the inverse operation of relative position mapping on the associated result to get the final set of target bounding boxes after the required tracking $\mathcal{T} $ to complete the tracking task.
\subsection{Relative Position Mapping Model}
\begin{figure}
  \centering
  \captionsetup{justification=centering}
  \includegraphics[width=0.45\textwidth]{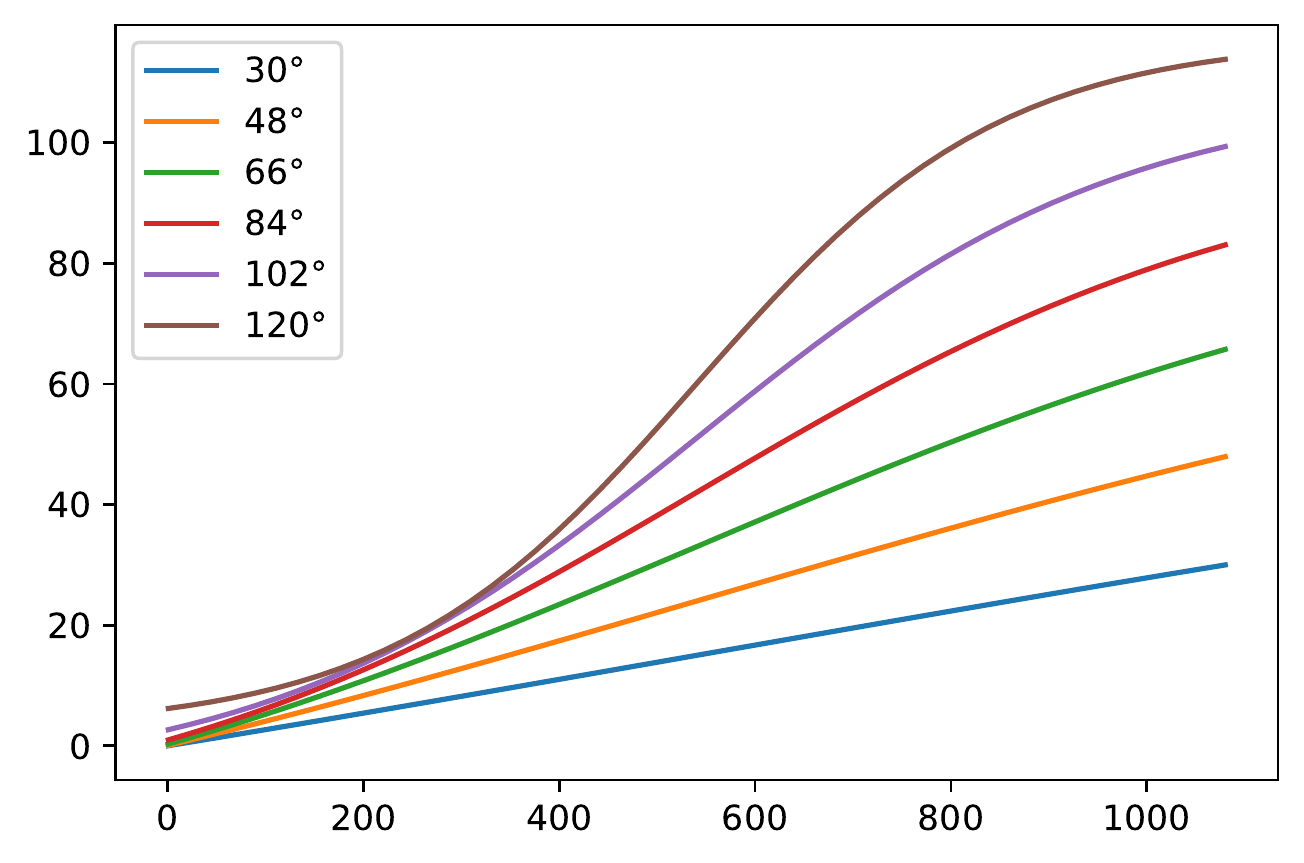}
  \caption{The relationship between $y$ and $\alpha_v $}
  \label{figure:4}
\end{figure}

We use the coordinates of the bounding box at the bottom of the middle to represent the position of the object in the picture, as indicated in the vertical view of Figure \ref{figure:2}, the relation of the vertical distance between the object's position and the image's lower limit. And the total image height corresponds to the target imaging angle $\alpha_v $ ratio and the camera's viewing angle $\beta_v $. The Equation \ref{eq:1} is the ratio between the size of the target imaging angle and $x,y$ is represented by $\mathcal{G} $. As the position of the object point in the video picture varies in the real world, the position of the point will also produce a nonlinear trend. This difference is generally considered to be the distortion produced by the video.
\begin{equation} \label{eq:1}
  \begin{split}
    \alpha_y = \mathcal{G}(y)
  \end{split}
\end{equation}
$y$ represents the distance between this target position on the video image and the bottom of the image, with a non-linear change in the imaging angle $\alpha_v $ as $y$ changes. The equation of the vertical mapping function $\mathcal{G}(y) $ is presented here:
\begin{equation} \label{eq:2}
  \mathcal{G}_v (y)=\left\{
  \begin{array}{lr}
    \dfrac{\beta_v }{2}-\tanh [(1-\dfrac{2y}{H}) \tan \dfrac{\beta_v }{2}]  ,  y<\dfrac{H}{2}\vspace{2ex}\\
    \dfrac{\beta_v }{2}+\tanh [(\dfrac{2y}{H}-1) \tan \dfrac{\beta_v }{2}]  ,  y \geqslant \dfrac{H}{2} \\
  \end{array}
  \right.
\end{equation}
Equation \ref{eq:2} in H is the total height of the video image, $ \beta_v $ is the size of the camera's vertical angle of vision, the camera's viewing angle depending on the focal length and factory configuration is often fixed in a surveillance video. We can also use other means, like deep learning algorithms, to get an approximate view. It can be seen from the formula in the vertical view of the camera that the size of the bounding box angle is related to the maximum vertical view of the camera and the height of the photo. As shown in Figure \ref{figure:4}, the horizontal coordinate indicates the position of the object spot on the image. The size of the flat coordinates is adjusted to 1080 here and the vertical coordinate denotes the angle of $\alpha_v $. The curves in each figure show the relation between the object's position and the target imaging angle at various maximum vertical viewing angle $\beta_v $. The bigger the camera's viewing angle, the more pronounced the non-linear change in the relationship between the position of the target image and the actual work.
For the horizontal mapping, $\alpha_h $ has the same calculation, as shown on the right side of Figure \ref{figure:2}, where the size of the pinch has to do with the width of the image $W$ and maximum view $\beta_h $ of horizontal camera, here is the formula:
\begin{equation} \label{eq:3}
  \mathcal{G}_h (x)=\left\{
  \begin{array}{lr}
    \dfrac{\beta_h }{2}-\tanh [(1-\dfrac{2x}{W}) \tan \dfrac{\beta_h }{2}]  ,x<\dfrac{W}{2}\vspace{2ex}           \\
    \dfrac{\beta_h }{2}+\tanh [(\dfrac{2x}{W}-1) \tan \dfrac{\beta_h }{2}]  ,x \geqslant \dfrac{W}{2} \\
  \end{array}
  \right.
\end{equation}
$x$ denotes the length of the horizontal coordinate of the target point location on the image. The mapping coefficient of the object is the ratio between the position of the target in the image and the position in the map plan. Because the position of the image will be distorted about the same work due to the perspective issue, the same target imaging angle response in the image position and the actual situation is different. as Figure \ref{figure:2} The ratio of $x^{'},y^{'}$ and $x,y$ is the target mapping coefficient $\varphi_h, \varphi_v $ as in Equation \ref{eq:4}, that directly reflects the effect of the image deformation on the actual relative position ratio of the target.
\begin{equation} \label{eq:4}
  \begin{split}
    \varphi_v = \dfrac{y^{'}}{y} = \mathcal{F}_v (\alpha_v , \beta_v , \gamma )\\
    \varphi_h = \dfrac{x^{'}}{x} = \mathcal{F}_h (\alpha_v ,\beta_v  \gamma )
  \end{split}
\end{equation}

Equation \ref{eq:4} is the mapping coefficient of the target in the horizontal and vertical directions, and it is possible to observe that the mapping coefficient is linked to the target imaging angle, maximum camera angle of view and camera grounding angle.
Figure \ref{figure:2} shows that in the vertical direction $y^{'}$ is calculated as follows.

\begin{equation} \label{eq:5}
  \begin{split}
    \theta& = 90\degree - \gamma - \dfrac{\beta }{2}
  \end{split}
\end{equation}
$\theta$ is the angle between the bottom edge of the camera view and the upright line. Assuming that the height of the camera above the ground is $h$ then $y^{'}$ is
\begin{equation} \label{eq:6}
  \begin{split}
    y^{'}& = h* [\tan(\alpha_v + \theta )-\tan\theta ]
  \end{split}
\end{equation}
The formula for $y$ is as follows.
\begin{equation} \label{eq:7}
  y=\left\{
  \begin{array}{lr}
    \dfrac{h*\tan\dfrac{\beta_v }{2} -\tan(\dfrac{\beta_v }{2} - \alpha_v ) }{\cos\theta },  \alpha_v < \dfrac{\beta_v }{2}\vspace{2ex}          \\
    \dfrac{h*\tan\dfrac{\beta_v }{2} +\tan(\alpha_v - \dfrac{\beta_v }{2}) }{\cos\theta },   \alpha_v \geqslant  \dfrac{\beta_v }{2} \\
  \end{array}
  \right.
\end{equation}

The formula for $\varphi_v $ is obtained as follows.
\begin{equation} \label{eq:8}
  \varphi_v =\left\{
  \begin{array}{lr}
    \dfrac{\cos\theta  * [ \tan(\alpha_v + \theta_v )-\tan\theta ]}{\tan\dfrac{\beta_v }{2} - \tan(\dfrac{\beta_v }{2} - \alpha_v )} , \alpha_v < \dfrac{\beta_v }{2} \vspace{2ex}       \\
    \dfrac{\cos\theta  * [ \tan(\alpha_v + \theta_v )-\tan\theta ]}{\tan\dfrac{\beta_v }{2} + \tan(\alpha_v - \dfrac{\beta_v }{2})} , \alpha_v \geqslant \dfrac{\beta_v }{2} \\
  \end{array}
  \right.
\end{equation}
The above equation represents the image deformation scale factor of the vertical coordinate of the target in the video image concerning the actual position in the vertical direction. The coefficient reflects the proportion of vertical deformation of the video image at a point.
\begin{figure}[h]
  \centering
  \captionsetup{justification=centering}
  \includegraphics[width=0.45\textwidth]{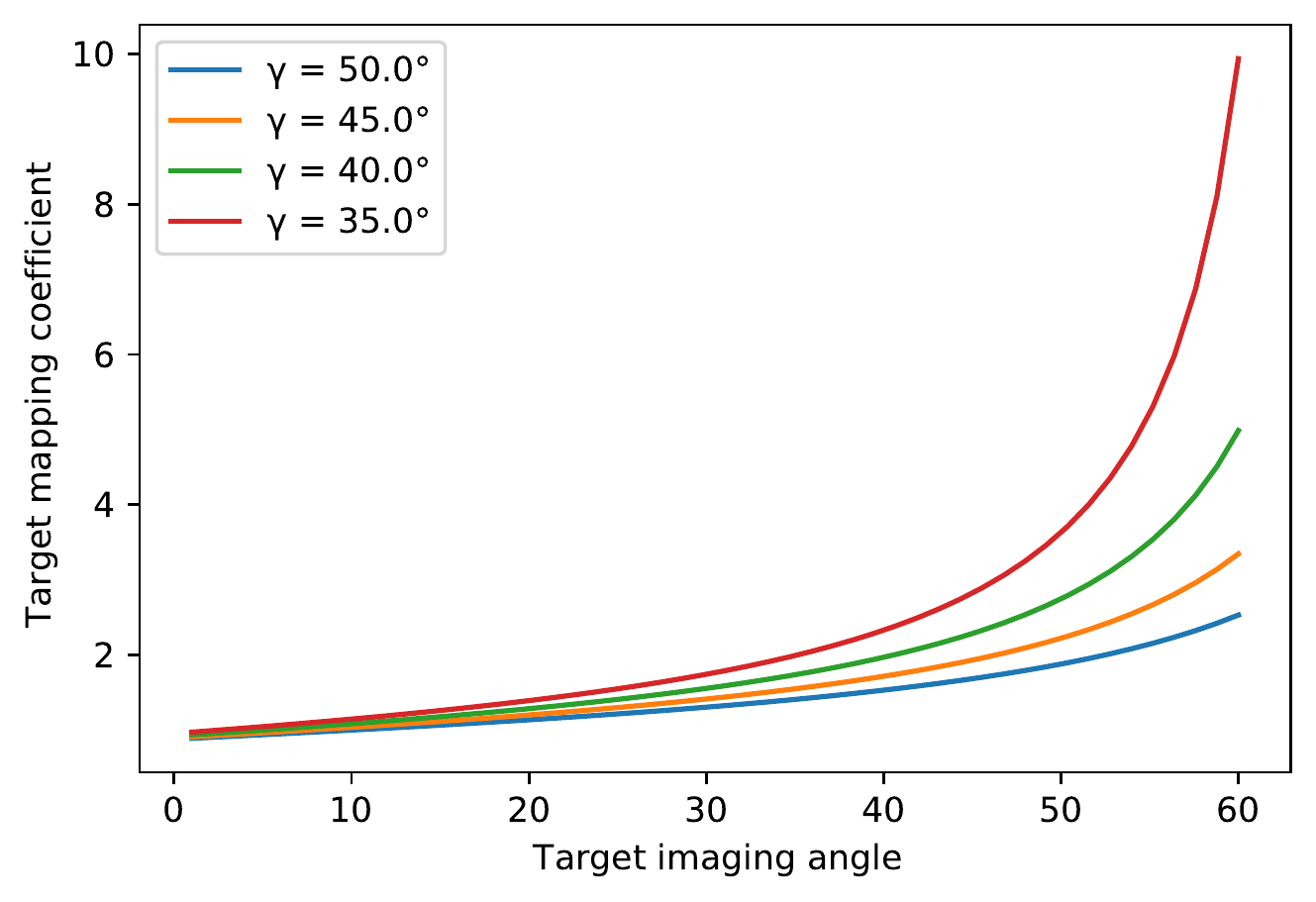}
  \caption{Relationship between camera-ground angle $\gamma $ and mapping coefficient $\varphi_v $ of vertical }
  \label{figure:5}
\end{figure}
Figure \ref{figure:5} is the schematic diagram of the relationship between the target in the vertical direction mapping coefficient $\varphi_v $ and the camera ground angle. It can be seen that when the camera ground angle is smaller, the vertical direction image mapping coefficient curve grows faster, which also fully reflects the deformation characteristics of the video image.
In terms of the lateral mapping coefficient, according to the schematic diagram on the right of Figure \ref{figure:2}, it can be seen that the lateral deformation generated by the increase of the vertical distance of the image also increases. The lateral mapping coefficient is expressed as shown in Equation \ref{eq:4}, and $\varphi_h $ is only related to the vertical imaging angle of the target and the angle between the camera and the ground. The following equation is available.
\begin{equation} \label{eq:9}
  \begin{split}
    \mathcal{F}_h(x) =  \dfrac{x^{'}}{x} = \dfrac{\cos\left\lvert \dfrac{\beta_v }{2} -\alpha_v \right\rvert * \cos\theta  }{ \cos\dfrac{\beta_v }{2}*\cos(\alpha_v +\theta ) }
  \end{split}
\end{equation}
$\theta$ is obtained for Equation \ref{eq:5} and $y^{'}$ is given in Equation \ref{eq:6}.
\begin{figure}[h]
  \centering
  \captionsetup{justification=centering}
  \includegraphics[width=0.45\textwidth]{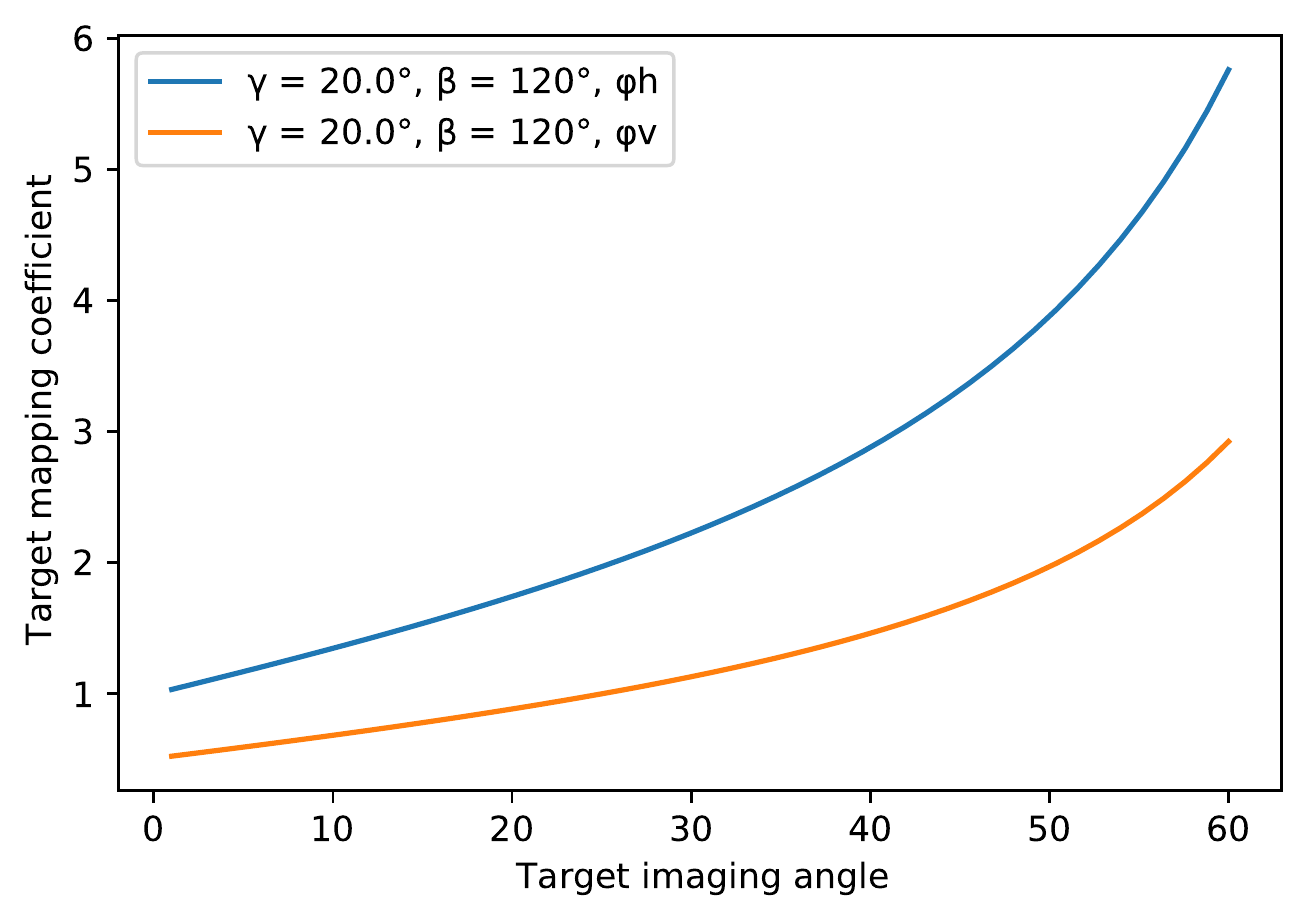}
  \caption{Schematic diagram of the relationship between horizontal mapping coefficient and vertical mapping coefficient }
  \label{figure:6}
\end{figure}
Figure \ref{figure:6} shows the comparison of $\varphi_v $ and $\varphi_h $ when $\beta_v = 120\degree, \gamma = 20\degree $. The figure shows that the video image mapping coefficient increases as the target imaging angle increases, and you may find that the lateral increase is greater than the vertical increase. Still, some are when $\beta_v $ is large enough, and as $\beta_v $ becomes smaller, the lateral mapping coefficient will gradually approach or be smaller than the steep increase. At this point, the mapped coordinates of the target video image position coordinates relative to the actual coordinates have been obtained, as shown below.
\begin{equation} \label{eq:10}
  \begin{split}
    \mathcal{R}_h(x) =& \dfrac{W}{2}(\varphi_{max} - \varphi_h )  + \mathcal{P}_{video}[x] * \varphi_h  \\
    \mathcal{R}_v(y) =&\mathcal{P}_{video}[y] * \varphi_v
  \end{split}
\end{equation}
This completes the mapping relationship from the video image to the actual ground.
\begin{figure*}[ht]
  \centering
  \captionsetup{justification=centering}
  \includegraphics[width=0.90\textwidth]{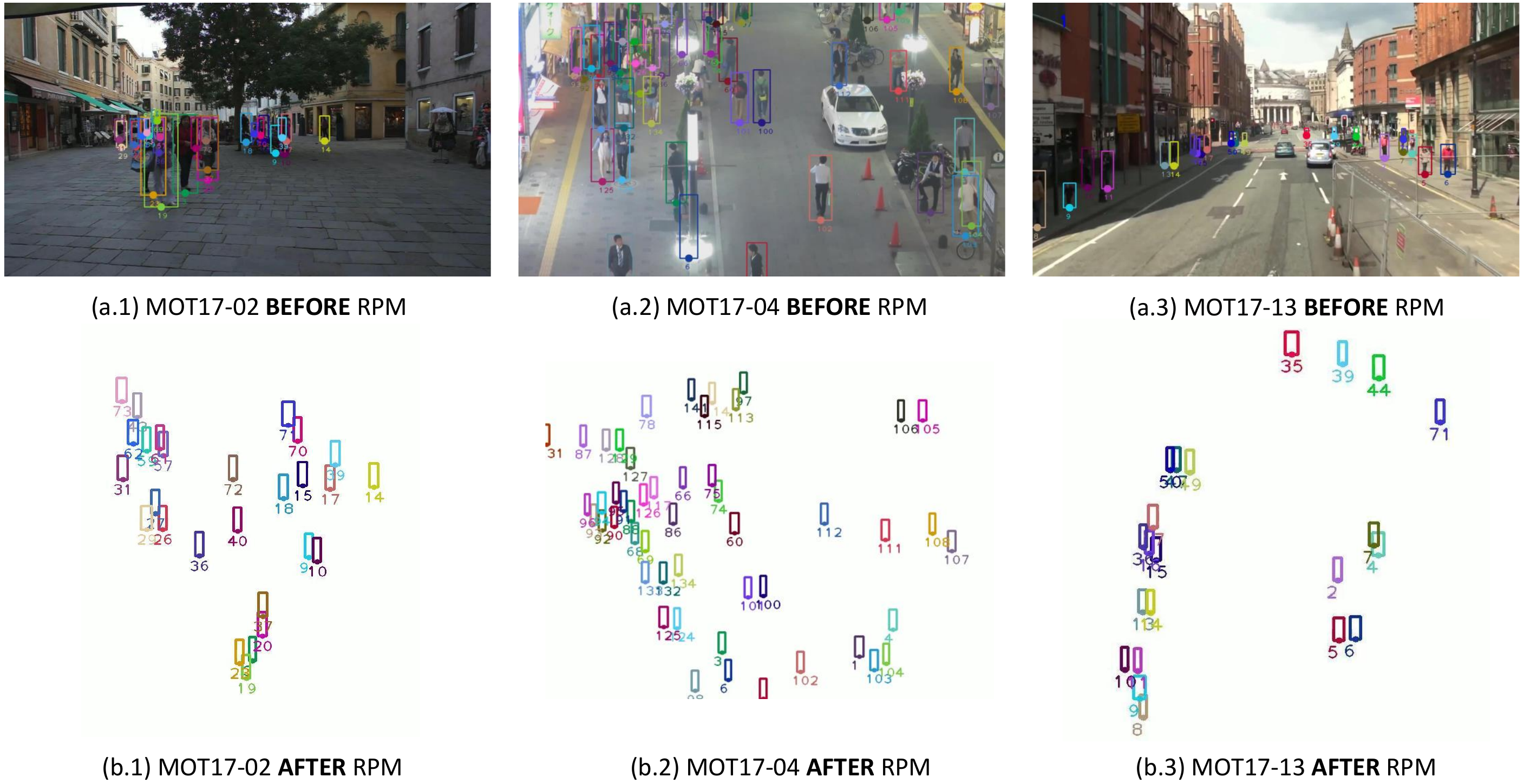}
  \caption{comparison of the image target labeling position of the dataset without RLM and the same frame target labeling position of the dataset after using RLM mapping}
  \label{figure:7}
\end{figure*}

\section{Extra Methods}
\subsection{Target Region Density}

Because different target regions exist in other scenes, we can focus more on potential targets in areas where targets exist than on the global picture. Similar to object detection's attention mechanism, we propose the TRD method for multi-object tracking. This method enables the multi-object tracking task to quickly perceive potential occlusion and interference areas. It allows a tracker to capture potentially missed tracking points more accurately and speedily by reducing the detection frame's filtering weight and tightening the normalization suggestion box. In addition, the TRD can assist in the determination of the normalization suggestion box. Since the region density parameter is updated in real-time, the normalization suggestion box can be more flexibly adapted to track the target for better tracking results.

\begin{figure}[h]
  \centering
  \captionsetup{justification=centering}
  \includegraphics[width=0.45\textwidth]{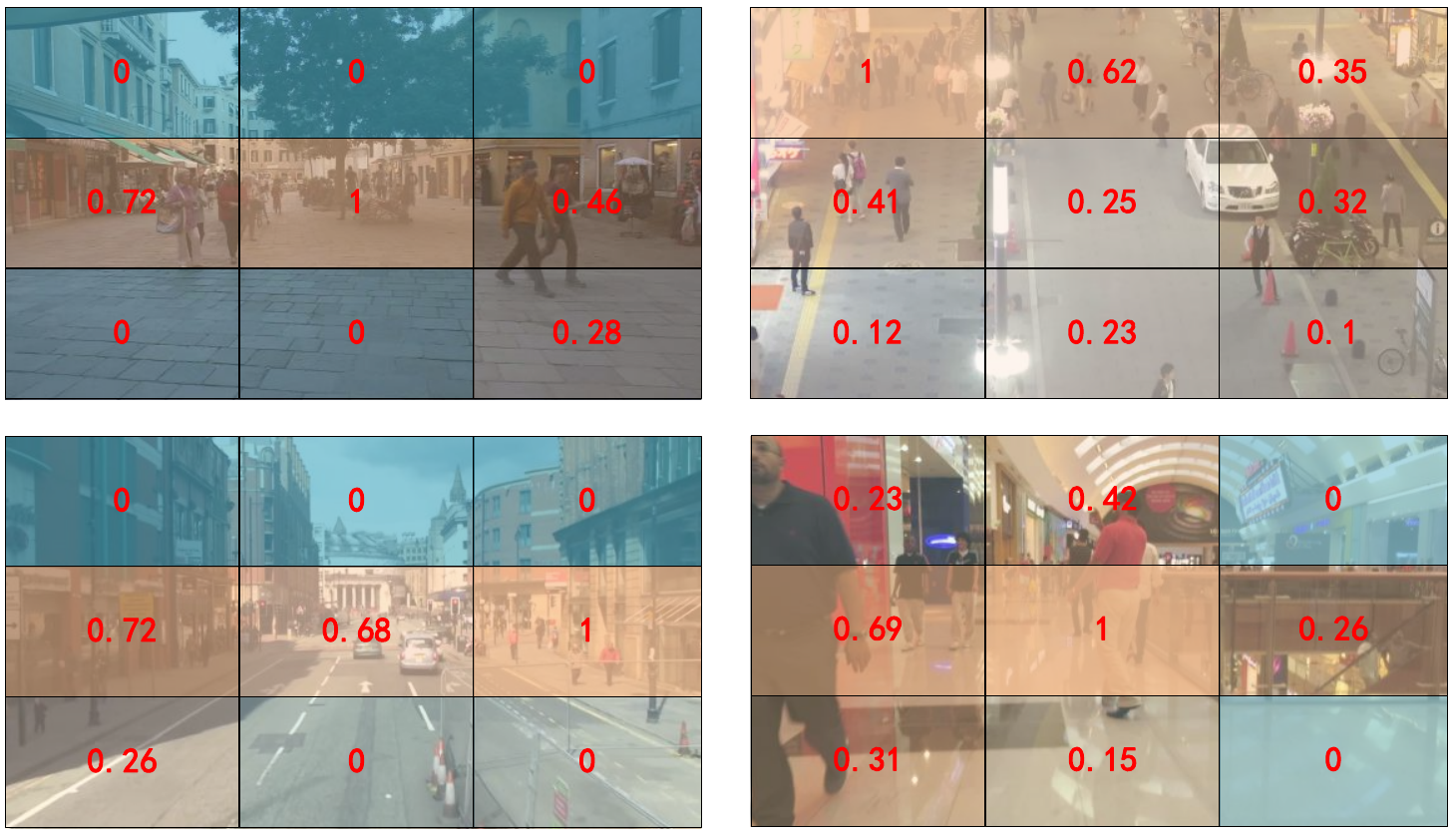}
  \caption{Diagram of target density matrix for different scenes}
  \label{figure:8}
\end{figure}

We divide the whole video image into nine regions; Each area is represented in figure \ref{figure:8}. In the review, the display can see the image of the highest density of the area. The computation of the density of each region is based on the object detection box located on the portion of each region for the weighted comparison and then the maximum region density as a reference point for the normalization process. By ascending recursive weights, different object detection boxes are added according to their location, size, and score. For instance, when the object passes through the region, the lower weight of the detection box will be significantly higher than the upper part of the weight. The TRD is calculated in the following manner:
\renewcommand\arraystretch{1}
\begin{equation}\label{arr:1}
  W=\left[
    \begin{array}{ccc}
      \rho_1 & \rho_2 & \rho_3 \\
      \rho_4 & \rho_5 & \rho_6 \\
      \rho_7 & \rho_8 & \rho_9 \\
    \end{array}
    \right],
  Q=\left[
    \begin{array}{ccc}
      1 & 1 & 1 \\
      1 & 1 & 1 \\
      1 & 1 & 1 \\
    \end{array}
    \right]
\end{equation}

\begin{equation} \label{eq:11}
  \rho_{max} = \max(\rho_n),  \:\:\:\:\: \rho_n \in W \\
\end{equation}

\begin{equation} \label{eq:12}
  \rho_n = \sum_{i=1}^n (Bbox_{i} \times Score_{i} \times \varpi )
\end{equation}

\begin{equation} \label{eq:13}
  A_{\rho } =  \prod_{i=1}^m[W *  (\frac{\rho_i}{\rho_{max}} * Q)]
\end{equation}

Equation \ref{eq:12}, $Bbox $ indicates the overlap between the original bounding box and the target region and the percentage of the original bounding box, $Score $ indicates the score of the original bounding box, $\varpi $ indicates the bottom-up weighted bounding box region percentage coefficient, $n$ shows the number of original bounding boxes in the target region, Equation Equation \ref{eq:13} $m$ denotes the number of target regions, $\rho _i$ represents the calculated density value of each region, $\rho _{max}$ indicates the maximum value of the calculated density of each region, $Q$ is the mask matrix. The normalized density matrix $A_\rho $ is obtained after multiplying with the region matrix $W$.
\begin{figure*}[htbp]
  \centering
  \captionsetup{justification=centering}
  \includegraphics[width=0.9\textwidth]{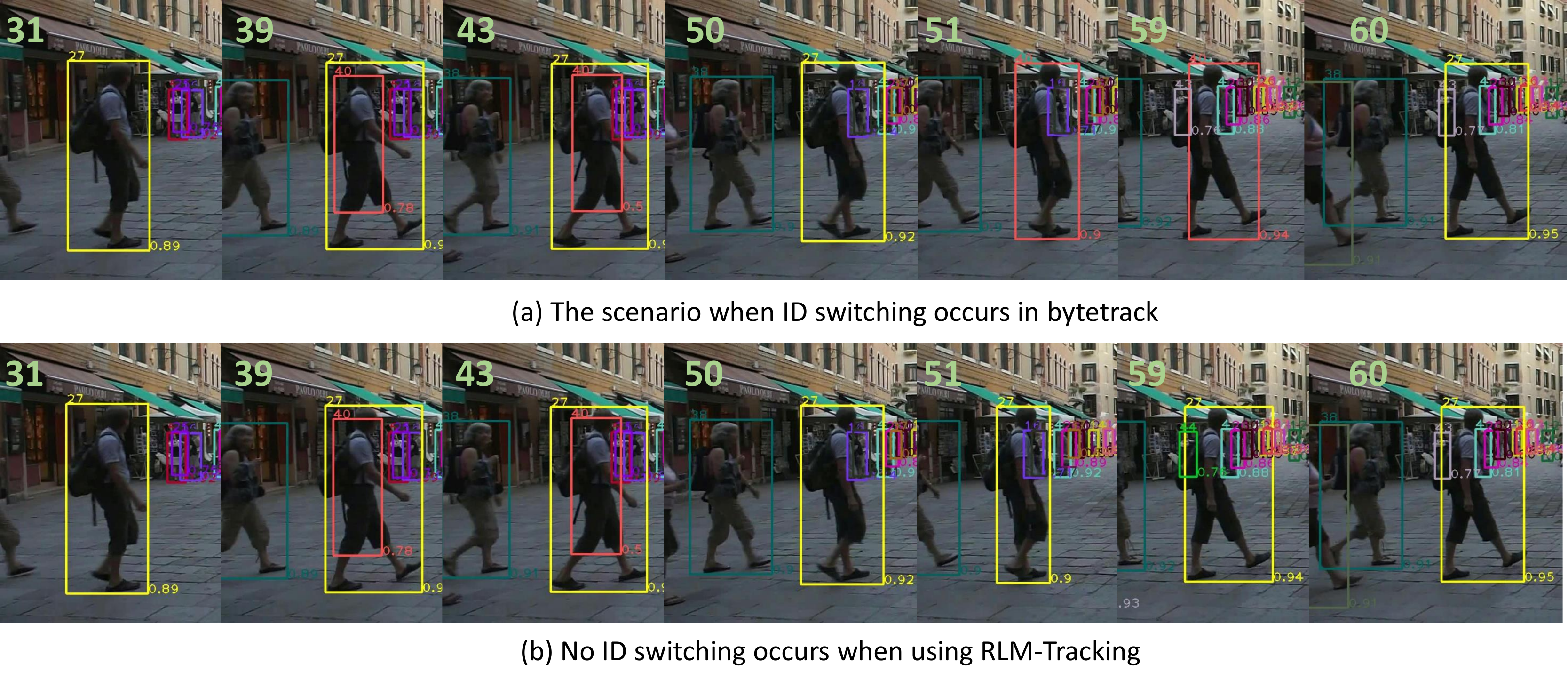}
  \caption{ID switching case due to false detection of bounding box overlap vs. using RLM correction}
  \label{figure:9}
\end{figure*}
\subsection{Normalized Bounding Box}
Once the relative position mapping operation is done, the mapping of the natural relations of the target is obtained for the distribution of critical points of the target’s position in the original ground-based video image. In the original image, the object’s depth is reflected by the size or area of the bounding box. However, a more unexpected ID switching occurs because of differences in morphological features, action behavior, and objects carried by the target in the scene. For example, an adult and a child within a short distance have a similar bounding box. Mapping the target’s relative position can genuinely reflect the physical relation between targets in the virtual environment without relying on the bounding box area. Consequently, the new mapping of the target location can use a fixed-width delimitation box to represent the situation of the target location. But the aspect ratio must be coherent with the original report to increase the distinction between the target classes. As shown in Figure \ref{figure:7}, the standard delimitation area of the mapped target projects the target into the original video image on the horizontal plane coordinate system, and the bounding box of different sizes within the original image is replaced by the same width delimitation box within the mapped image.

The mapping coefficient jointly determines the size of the mapped target normalized suggestion box $\varphi $ and the region density $A_{\rho } $, as well as the aspect ratio $\gamma $ of the original detected bounding box and the video frame rate $f$. To achieve the best intra-class recognition and avoid the ID switching caused by the overlapping normalization suggestion box, we set the normalization suggestion box at a maximum scale and minimum overlapping balance between full scale and minimum overlap.

\begin{equation} \label{eq:14}
  \begin{split}
    \mathcal{L}_{arg}=\frac{\sum_{i=1}^n(\mathcal{L}_i-\mathcal{L}_{i-1})}{n-1}
  \end{split}
\end{equation}
\begin{equation} \label{eq:15}
  \begin{split}
    W_i = \mathcal{L}_{arg} * (\frac{   A_{\varphi i}}{A_{\rho i} * f  c} ), \:\:\:\:A_{\rho i }>0,c\in \mathbb{R}
  \end{split}
\end{equation}
\begin{equation} \label{eq:16}
  \begin{split}
    S_{bbox} = \overline{W}  ^2 \gamma
  \end{split}
\end{equation}

The tracking effect can be enhanced by continuous tracking of image density by continuously optimizing the range of the standardized bounding box. Equation \ref{eq:15} shows the area of the standardized bounding box taken by continuous optimization of the density matrix, where $A_{\varphi i}$ denotes the mean value matrix of the target mapping coefficients in this recognition area, $\mathcal{L }_{arg} $ denotes the average spacing of the targets in the recognition area as in Equation Equation \ref{eq:14},$\overline{W} $ denotes the average obtained standardized width of the bounding box, $f$ is the video image frame rate, and $c$ is a frame rate control coefficient constant, it can be seen that the bounding box area can become smaller when the video frame rate is faster.

\section{Experiment}
\subsection{Configuration}

\textbf{dataset and backbone} \quad The MOT17\cite{milan2016mot16} and MOT20\cite{dendorfer2020mot20} datasets commonly used in the field of Multi-object tracking, what these two datasets contain both training and test datasets. Due to the limitations of submitting test data results, We divided the training dataset into two parts, the first half for the training set and the second half to verify the algorithm’s efficiency in object tracking. MOT17 dataset includes seven training videos and 7 test videos, with lengths between 7 and 90 seconds, frame rates between 14 and 30 fps, and resolutions of 1920$\times $1080 for most videos. The tracking dataset MOT20 has a more crowded and complex scene, with an average number of people in the image of about 170. Our training mixes the CrowdHuman, Cityperson, and ETHZ datasets for better detection outcomes for training as per the private detection protocol. To achieve good tracking results and a consistent validation environment, our task is based on the YOLOX\cite{ge2021yolox} object detection framework as the backbone network, which has excellent detection accuracy and speed and can better meet the demands of real-time detection for multi-object tracking tasks.

\textbf{evaluation metrics} \quad For the validation evaluation, we use the main evaluation measurements given by MOTChallenge. They are Multiple-Object Tracking Accuracy (MOTA), IDF1\cite{ristani2016performance } score, Higher-Order Tracking Accuracy (HOTA)\cite{luiten2021hota}, total number of false positives FP, detection accuracy DetA and other evaluation dimensions. This is done using the TrackEval tool\cite{luiten2020trackeval} to verify the assessment metrics.

\textbf{Implementation Details} \quad We uniformly use 8 NVIDIA Tesla V100S GPU graphics cards for training and parallel tests on the hardware and use them for testing on one of them, and the experimental code is implemented based on PyTorch\cite{pytorch2018pytorch}. Default $\tau_{high} $ threshold is set to 0.6. In contrast, the threshold $\tau_{low} $ is computed from a base of 0.3 on the regional probability density. We save 30 redundancy frames for the lost path to reappear, and the RLM tracking algorithm requires an initial evaluation of the camera variables as a function of the image content. Usually, target reconnaissance efforts in surveillance environments use fixed-angle surveillance camera systems to acquire videos captured regularly. From Equation \ref{eq:9}, we can see that the camera angle of view and the floor angle affect the relationship between the mapping position of the object in the image. In the experiment, we manually set the camera settings to fit the video scene's current situation.
\begin{equation} \label{eq:17}
  \begin{split}
    A =  \left[
      \begin{array}{cc}
        a_{11} & a_{12} \\
        a_{21} & a_{22} \\
      \end{array}
      \right],\quad
    B =  \left[
      \begin{array}{c}
        b_{1} \\
        b_{2} \\
      \end{array}
      \right]
  \end{split}
\end{equation}
\begin{equation} \label{eq:19}
  \begin{split}
    [x',y']^T = A\cdot [x,y]^T + B
  \end{split}
\end{equation}
\begin{equation} \label
  {eq:18}
  \begin{split}
    M_{k-1|k} = [A_{k-1|k}\quad B_{k-1|k}] = \left[
      \begin{array}{ccc}
        a_{11} & a_{12} & b_{1} \\
        a_{21} & a_{22} & b_{2} \\
      \end{array}
      \right]
  \end{split}
\end{equation}

\textbf{Preprocessing} \quad MOT17 and MOT20 data scenes with movement planes and unstable image data sets. The camera's angle of view does not change during photography with a portable camera, but the camera's angle to the ground changes slightly. It causes erratic changes in the object’s motion trajectory in the video, inter-image variations from camera motion also reduce tracking accuracy. Ensure continuous movement of the object, we use the OpenCV library \cite{bradski2000opencv} and vidgear\cite{vidgear} to process the video with real-time Affine Transformation, The original movement path of the object is preserved to the greatest extent possible. Equation \ref{eq:19} shows the affine transformation formula, where the $A$ matrix is used to represent the transformation of the image angle, and the $B$ matrix is used to represent the conversion of the translation distance of the image. The $M_{k-1|k}$ matrix in Equation \ref{eq:18} can be used to represent the transformation matrix from the $k-1$th frame to the $k$th frame in the video image.

\subsection{Verification Result}

We compared and rated the best methods and results for the tasks of the MOT17 and MOT20 multi-objective challenges using TrackEval, the official evaluation tool MOTChallenge.

\renewcommand\arraystretch{1.5}
\begin{table*}[htbp]
  \centering
  \begin{tabular}{l | c | c | c | c | c | c | c | c}
    \toprule
    Tracker& HOTA$(\uparrow)$ & MOTA$(\uparrow)$ & IDF1$(\uparrow)$ & IDs$(\downarrow)$ & DetA$(\uparrow)$ & AssA$(\uparrow)$ & SFDA$(\uparrow)$ & FPS$(\uparrow)$ \\
    \midrule
    SORT\cite{Bewley2016}      & 66.554           & 75.65            & 77.178           & 549               & 65.387           & 68.227           & 72.198           & \textbf{30}     \\
    DeepSORT\cite{Wojke2018}   & 66.28            & 76.736           & 77.595           & 515               & 65.636           & 67.462           & 72.437           & 11              \\
    MOTDT\cite{8486597}        & 66.806           & 77.057           & 78.347           & 413               & 65.93            & 68.156           & 72.865           & 13              \\
    BYTETrack\cite{Zhang2021a} & 67.733           & 77.888           & 79.928           & 384               & 66.663           & 69.345           & 73.422           & 29              \\
    \textbf{RLMTrack(ours)}    & \textbf{68.781}  & \textbf{78.044}  & \textbf{81.738}  & \textbf{378}      & \textbf{67.059}  & \textbf{70.984}  & \textbf{73.867}  & 27              \\
    \bottomrule
  \end{tabular}
  \caption{Comparison of evaluation indicators of different methods in MOT17 dataset}
  \label{label:1}
\end{table*}
\begin{table*}[htbp]
  \centering
  \begin{tabular}{l | c | c | c | c | c | c | c | c}
    \toprule
    Tracker& HOTA$(\uparrow)$ & MOTA$(\uparrow)$ & IDF1$(\uparrow)$ & IDs$(\downarrow)$ & DetA$(\uparrow)$ & AssA$(\uparrow)$ & SFDA$(\uparrow)$ & FPS$(\uparrow)$ \\
    \midrule
    SORT\cite{Bewley2016}      & 56.974           & 69.273           & 73.175           & 1990              & 58.022           & \textbf{56.112}  & 69.245           & \textbf{19}     \\
    DeepSORT\cite{Wojke2018}   & 53.816           & 70.331           & 67.599           & 2725              & 58.455           & 49.721           & 69.385           & 2               \\
    MOTDT\cite{8486597}        & 51.626           & 67.674           & 64.963           & 1876              & 56.3             & 47.487           & 68.344           & 4               \\
    BYTETrack\cite{Zhang2021a} & 56.155           & \textbf{71.931}  & 70.765           & 1565              & \textbf{59.374}  & 53.285           & \textbf{70.406}  & 17              \\
    \textbf{RLMTrack(ours)}    & \textbf{57.441}  & 71.864           & \textbf{73.267}  & \textbf{1524}     & 59.199           & 55.905           & 70.37            & 16              \\
    \bottomrule
  \end{tabular}
  \caption{Comparison of evaluation indicators of different methods in MOT20 dataset}
  \label{label:2}
\end{table*}
\textbf{MOT17 Dataset} \quad table \ref{label:1} in which we conducted a comparison experiment on several tracking models using the SORT algorithm in the MOT17 dataset VAL set, we can see that our method has improved the tracking accuracy of several algorithms in the MOT17 dataset to some extent and has improved performance in HOTA and IDF1 measurements. in MOTA and FPS metrics are similar in precision and speed to the BYTETrack method. This is because RLM typically uses fewer calculations and does not affect the model's operating efficiency. Like the figure \ref{figure:9}, the overlapping object fault detection box in ByteTrack may cause the ID to switch. In contrast, ID switching does not occur after using RLMT because the incorrect detection box is mapped further to the target, effectively reducing the number of ID switching.

\textbf{MOT20 Dataset} \quad In table \ref{label:2}, our method has considerably improved the HOTA and IDF1 and IDs metrics compared to several algorithms in the MOT20 dataset. The high-density of the MOT20 dataset causes some scrapped images to decrease DetA. However, such low-score structures increase the ability to switch IDs, and our algorithm rejects these images to get good tracking results.

\textbf{Limiting Case} \quad The association algorithm based on the Kalman filter relies mainly on the detection precision of the object detection image, which is subject to ID switching when object detection accuracy is low, or jitter is critical, which is a disadvantage of such detecting algorithms. We effectively reduce the number of ID switching in high-density regions in our experiments through the TRD method. This means some low-score but practical detection normalization suggestion boxes will be discarded. The method proposed in this paper is always the optimal solution under the comprehensive evaluation of both aspects of tracking speed to meet real-time requirements and have better tracking accuracy.

\subsection{Conclusion}

In this paper, we propose an enhanced multi-object tracking called RLM-Tracking, which uses a RLM model approach to project objects from the original video image to the map plane in the top view. and uses a TRD model to optimize bounding box filtering and generation of the normalization suggestion box to improve precision measurements for multi-object tracking. This technique can be applied to multi-object and other related fields and can be enhanced with a small overhead using this algorithm. Moreover, as mentioned in the paper, our work will continue to study the problem of determining the angle of two constant camera settings. We hope our research will contribute to the development of the MOT field.

\bibliography{ref.bib}

\end{document}